\documentclass[runningheads]{llncs}
\usepackage{graphicx}
\usepackage{hyperref}
\usepackage{subcaption}

\newcommand{\etal}{\textit{et al}. }

\title{DeepPrivacy: A Generative Adversarial Network for Face Anonymization}

\author{H\aa kon Hukkel\aa s \orcidID{0000-0001-9830-4931} \and
	Rudolf Mester \orcidID{0000-0002-6932-0606} \and
	Frank Lindseth \orcidID{0000-0002-4979-9218}}
\authorrunning{H. Hukkel\aa s et al.}
\institute{Department of Computer Science\\
	Norwegian University of Science and Technology\\
	Trondheim, Norway \\
	\email{\{hakon.hukkelas, rudolf.mester, frankl\}@ntnu.no}}
\begin{document}
\maketitle              
\begin{abstract}

We propose a novel architecture which is able to automatically anonymize faces in images while retaining the original data distribution. 
We ensure total anonymization of all faces in an image by generating images exclusively on privacy-safe information.
Our model is based on a conditional generative adversarial network, generating images considering the original pose and image background. The conditional information enables us to generate highly realistic faces with a seamless transition between the generated face and the existing background. 
Furthermore, we introduce a diverse dataset of human faces, including unconventional poses, occluded faces, and a vast variability in backgrounds. 
Finally, we present experimental results reflecting the capability of our model to anonymize images while preserving the data distribution, making the data suitable for further training of deep learning models. 
As far as we know, no other solution has been proposed that guarantees the anonymization of faces while generating realistic images.

\end{abstract}

\begin{figure}[h]
    \centering
    \includegraphics[width=1.0\textwidth]{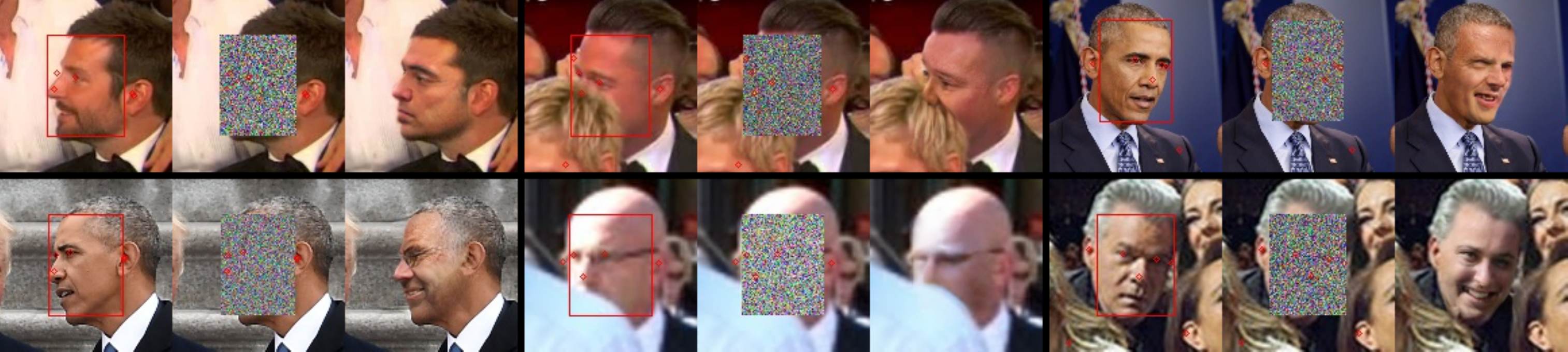}
    \caption{\textbf{DeepPrivacy Results} on a diverse set of images. The left image is the original image annotated with bounding box and keypoints, the middle image is the input image to our GAN, and the right image is the generated image. Note that our generator never sees any privacy-sensitive information.}
    \label{fig:annotated_results}
\end{figure}
\section{Introduction}

Privacy-preserving data-processing is becoming more critical every year; however, no suitable solution has been found to anonymize images without degrading the image quality. 
The General Data Protection Regulation (GDPR) came to effect as of 25th of May, 2018, affecting all processing of personal data across Europe.
GDPR requires regular consent from the individual for any use of their personal data. However, if the data  does not allow to identify an individual, companies are free to use the data without consent. To effectively anonymize images, we require a robust model to replace the original face, without destroying the existing data distribution; that is: the output should be a realistic face fitting the given situation. 

Anonymizing images, while retaining the original distribution, is a challenging task. The model is required to remove all privacy-sensitive information, generate a highly realistic face, and the transition between original and anonymized parts has to be seamless. 
This requires a model that can perform complex semantic reasoning to generate  a new anonymized face.
For practical use, we desire the model to be able to manage a broad diversity of images, poses, backgrounds, and different persons. Our proposed solution can successfully anonymize images in a large variety of cases, and create realistic faces to the given conditional information. 

Our proposed model, called \textit{DeepPrivacy}, is a conditional generative adversarial  network \cite{Goodfellow2014,Mirza2014}. Our generator considers the existing background and a sparse pose annotation to generate realistic anonymized faces. 
The generator has a U-net architecture \cite{ronneberger2015u} that generates images with a resolution of $128 \times 128$. 
The model is trained with a progressive growing training technique \cite{Karras2017} from a starting resolution of $8 \times 8$ to $128 \times 128$, which substantially improves the final image quality and overall training time. 
By design, our generator never observes the original face, ensuring removal of any privacy-sensitive information.

For practical use, we assume no demanding requirements for the object and keypoint detection methods. Our model requires two simple annotations of the face: (1) a bounding box annotation to identify the privacy-sensitive area, and (2) a sparse pose estimation of the face, containing keypoints for the ears, eyes, nose, and shoulders; in total seven keypoints. This keypoint annotation is identical to what Mask R-CNN \cite{He2017} provides.

We provide a new dataset of human faces, \textit{Flickr Diverse Faces} (FDF), which consists of 1.47M faces with a bounding box and keypoint annotation for each face. This dataset covers a considerably large diversity of facial poses, partial occlusions, complex backgrounds, and different persons. We will make this dataset publicly available along with our source code and pre-trained networks\footnote{Code: \url{www.github.com/hukkelas/DeepPrivacy}}\footnote{FDF Dataset: \url{www.github.com/hukkelas/FDF}}.

We evaluate our model by performing an extensive qualitative and quantitative study of the model's  ability to retain the original data distribution. 
We anonymize the validation set of the WIDER-Face dataset \cite{yang2016wider}, then run face detection on the anonymized images to measure the impact of anonymization on Average Precision (AP). 
DSFD \cite{li2018dsfd} achieves $99.3\%$ ($95.9\%$ out of $96.6\%$ AP), $99.3\%$ ($95.0\%/95.7\%$), and $99.3\%$ ($89.8\%/90.4\%$) of the original AP on the easy, medium, and hard difficulty, respectively.
On average, it achieves $99.3\%$ of the original AP.
In contrast, traditional anonymization techniques, such as $8x8$ pixelation achieves $96.7\%$, heavy blur $90.5\%$, and black-out $41.4\%$ of the original performance.
Additionally, we present several ablation experiments that reflect the importance of a large model size and conditional pose information to generate high-quality faces.

In summary, we make the following contributions:
\begin{itemize}

    \item We propose a novel generator architecture to anonymize faces, which ensures 100\% removal of privacy-sensitive information in the original face. The generator can generate realistic looking faces that have a seamless transition to the existing background for various sets of poses and contexts.
    \item We provide the FDF dataset, including 1.47M faces with a tight bounding box and keypoint annotation for each face. The dataset covers a considerably larger diversity of faces compared to previous datasets.
    

\end{itemize}

\section{Related Work}

\textbf{De-Identifying Faces:}
Currently, there exists a limited number of  research studies on the task of removing privacy-sensitive information from an image including a face. 
Typically, the approach chosen is to alter the original image such that we remove all the privacy-sensitive information. 
These methods can be applied to all images; however, there is no assurance that these methods remove all privacy-sensitive information. 
Naive methods that apply simple image distortion have been discussed numerous times in literature \cite{Boyle2000ThePrivacy,Neustaedter2006BlurConferencing,Gross2009FaceDe-identification,Newton2005,Gross2006Model-basedDe-identification}, such as pixelation and blurring; but, they are inadequate for removing the privacy-sensitive information \cite{Gross2006Model-basedDe-identification,Neustaedter2006BlurConferencing,Newton2005}, and they alter the data distribution substantially.

K-same family of algorithms \cite{Gross2006Model-basedDe-identification,Jourabloo2015AttributeDe-identification,Newton2005} implements the k-anonymity algorithm \cite{Sweeney2002K-anonymity:Privacy} for face images. Newton \etal prove that the k-same algorithm can remove all privacy-sensitive information; but, the resulting images often contain "ghosting" artifacts due to small alignment errors \cite{Gross2006Model-basedDe-identification}.

Jourabloo \etal \cite{Jourabloo2015AttributeDe-identification} look at the task of de-identification grayscale images while preserving a large set of facial attributes. This is different from our work, as we do not directly train our generative model to generate faces with similar attributes to the original image. In contrast, our model is able to perform complex semantic reasoning to generate a face that is coherent with the overall context information given to the network, yielding a highly realistic face.

\textbf{Generative Adversarial Networks} (GANs) \cite{Goodfellow2014} is a highly successful training architecture to model a natural image distribution. GANs enables us to  generate new images, often indistinguishable from the real data distribution. It has a broad diversity of application areas, from general image generation \cite{Brock2018,Karras2017,Karras2018ANetworks,Zhang2018}, text-to-photo generation \cite{Zhang2017}, style transfer \cite{Huang2017ArbitraryNormalization,Ruder2016} and much more. With the numerous contributions since its conception, it has gone from a beautiful theoretical idea to a tool we can apply for practical use cases. 
In our work, we show that GANs are an efficient tool to remove privacy-sensitive information without destroying the original image quality.

 Ren \etal \cite{Ren2018} look at the task of anonymizing video data by using GANs. They perform anonymization by altering each pixel in the original image to hide the identity of the individuals. In contrast to their method, we can ensure the removal of all privacy-sensitive information, as our generative model never observes the original face.

\textbf{Progressive Growing of GANs} \cite{Karras2017} propose a novel training technique to generate faces progressively, starting from a resolution of $4x4$ and step-wise increasing it to $1024x1024$. This training technique improves the final image quality and overall training time. Our proposed model uses the same training technique; however, we perform several alterations to their original model to convert it to a conditional GAN. With these alterations, we can include conditional information about the context and pose of the face.
Our final generator architecture is similar to the one proposed by Isola \etal \cite{Isola2017}, but we introduce conditional information in several stages.

\textbf{Image Inpainting} is a closely related task to what we are trying to solve, and it is a widely researched area for generative models \cite{Jo2019SC-FEGAN:Color,Li2017,Liu2018,Yeh2017}.
Several research studies have looked at the task of face completion with a generative adversarial network \cite{Li2017,Yeh2017}. They mask a specific part of the face and try to complete this part with the conditional information given.
From our knowledge, and the qualitative experiments they present in their papers, they are not able to mask a large enough section to remove all privacy-sensitive information. As the masked region grows, it requires a more advanced generative model that understands complex semantic reasoning, making the task considerably harder. 
Also, their experiments are based on the Celeb-A dataset \cite{Liu2018}, primarily consisting of celebrities with low diversity in facial pose, making models trained on this dataset unsuitable for real-world applications. 



\section{The Flickr Diverse Faces Dataset}

\begin{figure}	
	\centering
	\includegraphics[width=1.0\textwidth]{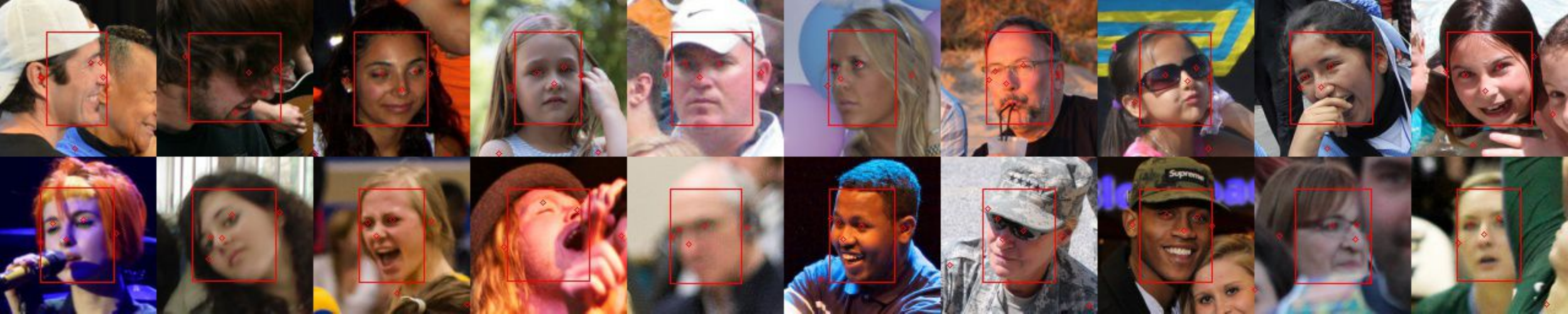}
	\caption[Flickr Diverse Datset Examples]{\textbf{The FDF dataset}. Each image has a sparse keypoint annotation (7 keypoints) of the face and a tight bounding box annotation. We recommend the reader to zoom in.}
	\label{fig:fdf_examples}
\end{figure}

\textit{FDF} (Flickr Diverse Faces) is a new dataset of human faces, crawled from the YFCC-100M dataset \cite{Thomee2015YFCC100M:Research}. It consists of 1.47M human faces with a minimum resolution of $128 \times 128$, containing facial keypoints and a bounding box annotation for each face. The dataset has a vast diversity in terms of age, ethnicity, facial pose, image background, and face occlusion. 
Randomly picked examples from the dataset can be seen in \autoref{fig:fdf_examples}.
The dataset is extracted from scenes related to traffic, sports events, and outside activities.
In comparison to the FFHQ \cite{Karras2018ANetworks} and Celeb-A \cite{Liu2018} datasets, our dataset is more diverse in facial poses and it contains significantly more faces; however, the FFHQ dataset has a higher resolution.

The FDF dataset is a high-quality dataset with few annotation errors. 
The faces are automatically labeled with state-of-the-art keypoint and bounding box models, and we use a high confidence threshold for both the keypoint and bounding box predictions.
The faces are  extracted from $1.08M$ images in the YFCC100-M dataset.  For keypoint estimation, we use Mask R-CNN \cite{He2017}, with a ResNet-50 FPN backbone \cite{Lin2017FeatureDetection}. For bounding box annotation, we use the Single Shot Scale-invariant Face Detector \cite{Zhang2017S3fd:Detector}. 
To combine the predictions, we match a keypoint with a face bounding box if the eye and nose annotation are within the bounding box.
Each bounding box and keypoint has a single match, and we match them with a greedy approach based on descending prediction confidence.
\section{Model}

Our proposed model is a conditional GAN, generating images based on the surrounding of the face and sparse pose information. \autoref{fig:annotated_results} shows the conditional information given to our network, and Appendix A has a detailed description of the pre-processing steps.
We base our model on the one proposed by Karras \etal \cite{Karras2017}. Their model is a non-conditional GAN, and we perform several alterations to include conditional information.

We use seven keypoints to describe the pose of the face: left/right eye, left/right ear, left/right shoulder, and nose. To reduce the number of parameters in the network, we pre-process the pose information into a one-hot encoded image of size $K \times M \times M$, where $K$ is the number of keypoints and $M$ is the target resolution. 

Progressive growing training technique is crucial for our model's success.
We apply progressive growing to both the generator and discriminator to grow the networks from a starting resolution of $8$. We double the resolution each time we expand our network until we reach the final resolution of $128 \times128$. 
The pose information is included for each resolution in the generator and discriminator, making the pose information finer for each increase in resolution.

\begin{figure}
    \centering
    \includegraphics[width=0.95\textwidth]{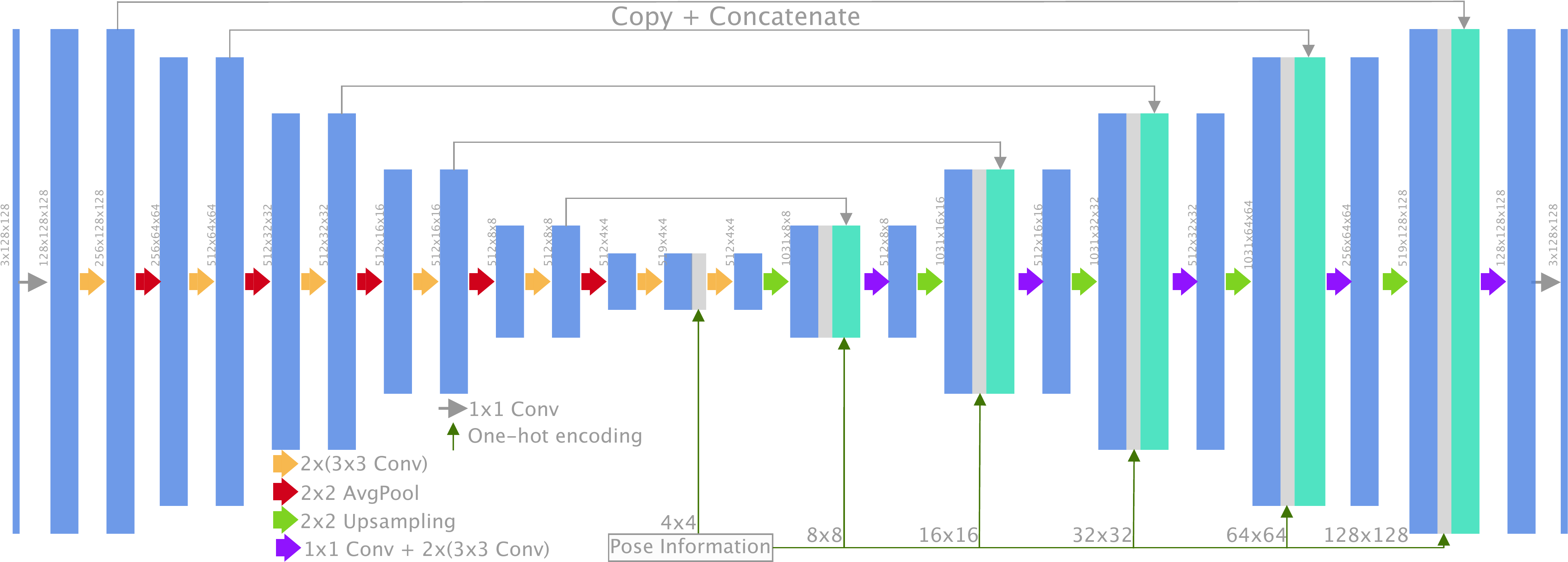}
    \caption{\textbf{Generator Architecture} for $128 \times 128$ resolution. Each convolutional layer is followed by pixel normalization \cite{Karras2017} and LeakyReLU($\alpha = 0.2$).
    After each upsampling layer, we concatenate the upsampled output with pose information and the corresponding skip connection.}
    \label{fig:generator_architecture}
\end{figure}

\subsection{Generator Architecture}

\autoref{fig:generator_architecture} shows our proposed generator architecture for $128 \times 128$ resolution. Our generator has a U-net \cite{ronneberger2015u} architecture to include background information.
The encoder and decoder have the same number of filters in each convolution, but the decoder has an additional $1\times 1$ bottleneck convolution after each skip connection. This bottleneck design reduces the number of parameters in the decoder significantly. To include the pose information for each resolution, we concatenate the output after each upsampling layer with pose information and the corresponding skip connection. The general layer structure is identical to Karras \etal \cite{Karras2017}, where we use pixel replication for upsampling, pixel normalization and LeakyReLU after each convolution, and equalized learning rate instead of careful weight initialization. 

\textbf{Progressive Growing:} Each time we increase the resolution of the generator, we add two $3 \times 3$ convolutions to the start of the encoder and the end of the decoder. We use a transition phase identical to Karras \etal \cite{Karras2017} for both of these new blocks, making the network stable throughout training. We note that the network is still unstable during the transition phase, but it is significantly better compared to training without progressive growing.

\subsection{Discriminator Architecture}

Our proposed discriminator architecture is identical to the one proposed by Karras \etal \cite{Karras2017}, with a few exceptions. First, we include the background information as conditional input to the start of the discriminator, making the input image have six channels instead of three. Secondly, we include pose information at each resolution of the discriminator. The pose information is concatenated with the output of each downsampling layer, similar to the decoder in the generator. Finally, we remove the mini-batch standard deviation layer presented by Karras \etal \cite{Karras2017}, as we find the diversity of our generated faces satisfactory.

The adjustments made to the generator doubles the number of total parameters in the network. To follow the design lines of Karras \etal \cite{Karras2017}, we desire that the complexity in terms of the number of parameters to be similar for the discriminator and generator. We evaluate two different discriminator models, which we will name the \textit{deep discriminator} and the \textit{wide discriminator}. The deep discriminator doubles the number of convolutional layers for each resolution. To mimic the skip-connections in the generator, we wrap the convolutions for each resolution in residual blocks.
The wider discriminator keeps the same architecture; however, we increase the number of filters in each convolutional layer by a factor of $\sqrt{2}$. 
\section{Experiments}

\begin{figure}
    \centering
    \includegraphics[width=1.0\textwidth]{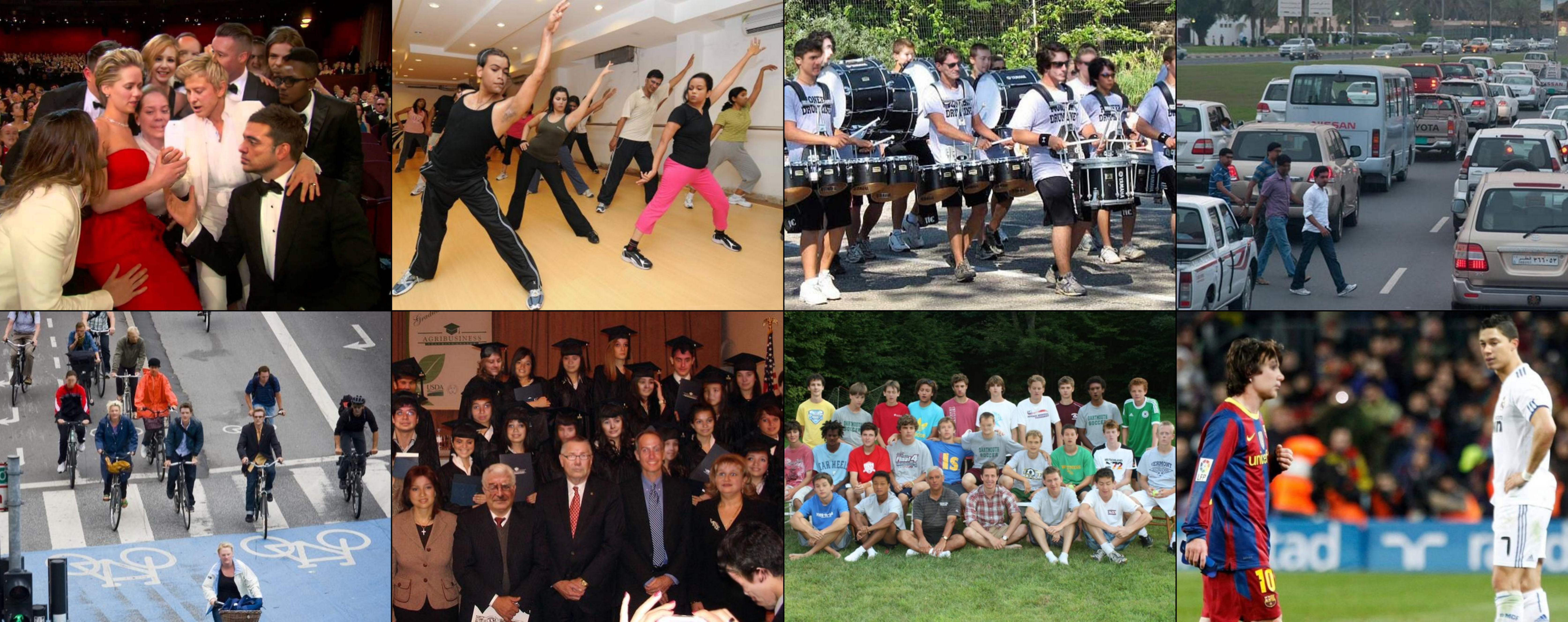}
    \caption{\textbf{Anonymized Images from DeepPrivacy}. Every single face in the images has been generated. We recommend the reader to zoom in.}
    \label{fig:more_results}
\end{figure}

DeepPrivacy can robustly generate anonymized faces for a vast diversity of poses, backgrounds, and different persons.
From qualitative evaluations of our generated results on  the WIDER-Face dataset \cite{yang2016wider}, we find our proposed solution to  be robust to a broad diversity of images.
\autoref{fig:more_results} shows several results of our proposed solution on the WIDER-Face dataset. Note that the network is trained on the FDF dataset; we do not train on any images in the WIDER-Face dataset.

We evaluate the impact of anonymization on the WIDER-Face \cite{yang2016wider} dataset.
We measure the AP of a face detection model on the anonymized dataset and compare this to the original dataset.
We report the standard metrics for the different difficulties for WIDER-Face. Additionally, we perform several ablation experiments on our proposed FDF dataset.

Our final model is trained for 17 days, 40M images, until we observe no qualitative differences between consecutive training iterations. It converges to a Fr\`echect Inception Distance (FID) \cite{heusel2017gans} of $1.53$. Specific training details and input pre-processing  are given in Appendix A.

\subsection{Effect of Anonymization for Face Detection}

\begin{table}
	\centering
	\caption[Average Precision on WIDER-Face validation set]{\textbf{Face Detection AP} on the WIDER Face \cite{yang2016wider} validation dataset. The face detection method used is DSFD \cite{li2018dsfd}, the current state-of-the-art on WIDER-Face. }
	\begin{tabular}{l|ccc}
		\hline
		Anonymization method & Easy & Medium & Hard \\
		\hline
		No Anonymization \cite{li2018dsfd} & $96.6\%$ & $95.7\%$  & $90.4\%$ \\
		\hline
		
		Blacked out & $24.9\%$ & $36.3\%$  & $54.8\%$ \\
		Pixelation ($16x16$) & 95.3\% & 94.9\% & \textbf{90.2\%} \\ 
		Pixelation ($8x8$) & 91.4\% & $92.3\%$ & $88.9\%$ \\ 
		9x9 Gaussian Blur ($\sigma = 3)$ & $95.3\%$ & $92.8\%$  & $84.7\%$ \\
		Heavy Blur (filter size = 30\% face width) & $83.4\%$ & $86.3\%$  & $86.1\%$ \\
		\textbf{DeepPrivacy} (Ours) & \textbf{95.9\%} & \textbf{95.0\%} & 89.8\% \\
		\hline
		
	\end{tabular}
	\vspace{2mm}

	\label{tab:wider_face_results}
\end{table}

\begin{figure}
    \centering
    \includegraphics[width=.99\textwidth]{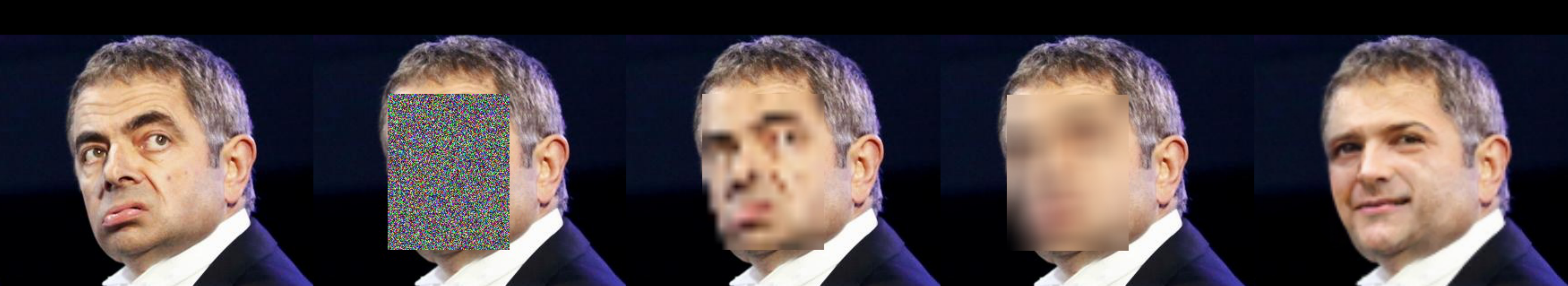}
    \caption{\textbf{Different Anonymization Methods} on a face in the WIDER Face validation set.}
    \label{fig:anonymization_methods}
\end{figure}

\autoref{tab:wider_face_results} shows the AP of different anonymization techniques on the WIDER-Face validation set.
In comparison to the original dataset, DeepPrivacy only degrades the AP by $0.7\%$, $0.7\%$, and $0.6\%$ on the easy, medium, and hard difficulties, respectively.


We compare DeepPrivacy anonymization to simpler anonymization methods; black-out, pixelation, and blurring. \autoref{fig:anonymization_methods} illustrates the different anonymization methods. 
DeepPrivacy generally achieves a higher AP compared to all other methods, with the exception of $16 \times 16$ pixelation.

Note that $16 \times 16$ pixelation does not affect a majority of the faces in the dataset. 
For the "hard" challenge, $0\%$ of the faces has a resolution larger than $16 \times 16$.
For the easy and medium challenge, $43\%$ and $29.9\%$ has a resolution larger than $16 \times 16$.
The observant reader might notice that for the "hard" challenge, $16 \times 16$ pixelation should have no effect; however, the AP is degraded in comparison to the original dataset (see \autoref{tab:wider_face_results}).
We believe that the AP on the "hard" challenge is degraded due to anonymizing faces in easy/medium challenge can affect the model in cases where faces from "hard" and easy/medium are present in the same image.

\textbf{Experiment Details:} For the face detector we use the current state-of-the-art, Dual Shot Face Detector (DSFD) \cite{li2018dsfd}.
The WIDER-Face dataset has no facial keypoint annotations; therefore,
we automatically detect keypoints for each face with the same method as used for the FDF dataset.
To match keypoints with a bounding box, we use the same greedy approach as earlier. 
Mask R-CNN \cite{He2017} is not able to detect keypoints for all faces, especially in cases with high occlusion, low resolution, or faces turned away from the camera.
Thus, we are only able to anonymize $43\%$ of the faces in the validation set.
Of the faces that are not anonymized, $22\%$ are partially occluded, and $30\%$ are heavily occluded. For the remaining non-anonymized faces, $70\%$ has a resolution smaller than $14x14$. Note that for each experiment in \autoref{tab:wider_face_results}, we anonymize the same bounding boxes.

\subsection{Ablation Experiments}

We perform several ablation experiments to evaluate the model architecture choices. We report the Fr\`echet Inception Distance \cite{heusel2017gans} between the original images and the anonymized images for each experiment. 
We calculate FID from a validation set of $50,000$ faces from the FDF dataset.
The results are shown in \autoref{tab:ablation_experiments} and discussed in detail next.

\begin{table}
	\caption[Ablation Experiments Results]{\textbf{Ablation Experiments} with our model. We report the Fr\`echet Inception Distance (FID) on the FDF validation dataset, after showing the discriminator $30.0M$ images (lower is better). For results in \autoref{tab:ablation_pose} and \autoref{tab:ablation_discriminator}, we use a model size of $12M$ parameters for both the generator and discriminator. 
	*Reported after $20.0M$ images, as the deep discriminator diverged after this.}
	\centering
	\begin{subtable}[h]{0.25\textwidth}
		\captionsetup{justification=centering}
		
		\caption{Result of using conditional pose.}
		\centering
		
		\begin{tabular}{c|c}
			\hline
			Model & FID\\
			\hline
			With Pose & \textbf{2.71} \\ 
			Without Pose & 3.36 \\
			\hline
		\end{tabular}
		
		\label{tab:ablation_pose}
	\end{subtable}
	\hfill
	\begin{subtable}[h]{0.45\textwidth}
		\captionsetup{justification=centering}
		\caption{Result of the deep and wide discriminator.}
		\centering
		
		\begin{tabular}{c|c}
			\hline
			Discriminator & FID \\ 
			\hline
			Deep Discriminator* & $9.327$\\ 
			Wide Discriminator* & \textbf{3.86}\\ 
			\hline
		\end{tabular}
		
		\label{tab:ablation_discriminator}
	\end{subtable}
	\hfill
	\begin{subtable}[h]{0.25\textwidth}
		\centering
		\captionsetup{justification=centering}
		\caption{Result of different model sizes.}
		\begin{tabular}{c|c}
			\hline
			\#parameters & FID \\ 
			\hline
			12M & $2.71$\\ 
			46M & \textbf{1.84} \\ 
			\hline
		\end{tabular}
		
		\label{tab:ablation_parameters}
	\end{subtable}
	\vspace{2mm}

	\label{tab:ablation_experiments}
\end{table}

\textbf{Effect of Pose Information:}
Pose of the face provided as conditional information improves our model significantly, as seen in \autoref{tab:ablation_pose}. The FDF dataset has a large variance of faces in different poses, and we find it necessary to include sparse pose information to generate realistic faces.
In contrast, when trained on the Celeb-A dataset, our model completely ignores the given pose information.

\textbf{Discriminator Architecture:}
\autoref{tab:ablation_discriminator} compares the quality of images for a deep and wide discriminator. With a deeper network, the discriminator struggles to converge, leading to poor results. We use no normalization layers in the discriminator, causing deeper networks to suffer from exploding forward passes and vanishing gradients.
Even though, Brock \etal \cite{Brock2018} also observe  similar results; a deeper network architecture degrades the overall image quality.
Note that we also experimented with a discriminator with no modifications to number of parameters, but this was not able to generate realistic faces.

\textbf{Model Size:}
We empirically observe that increasing the number of filters in each convolution improves image quality drastically. 
As seen in \autoref{tab:ablation_parameters}, we train two models with $12M$ and $46M$ parameters. Unquestionably, increasing the number of parameters generally improves the image quality. For both experiments, we use the same hyperparameters; the only thing changed is the number of filters in each convolution. 


\section{Limitations}

\begin{figure}
    \centering
    \includegraphics[width=1.0\textwidth]{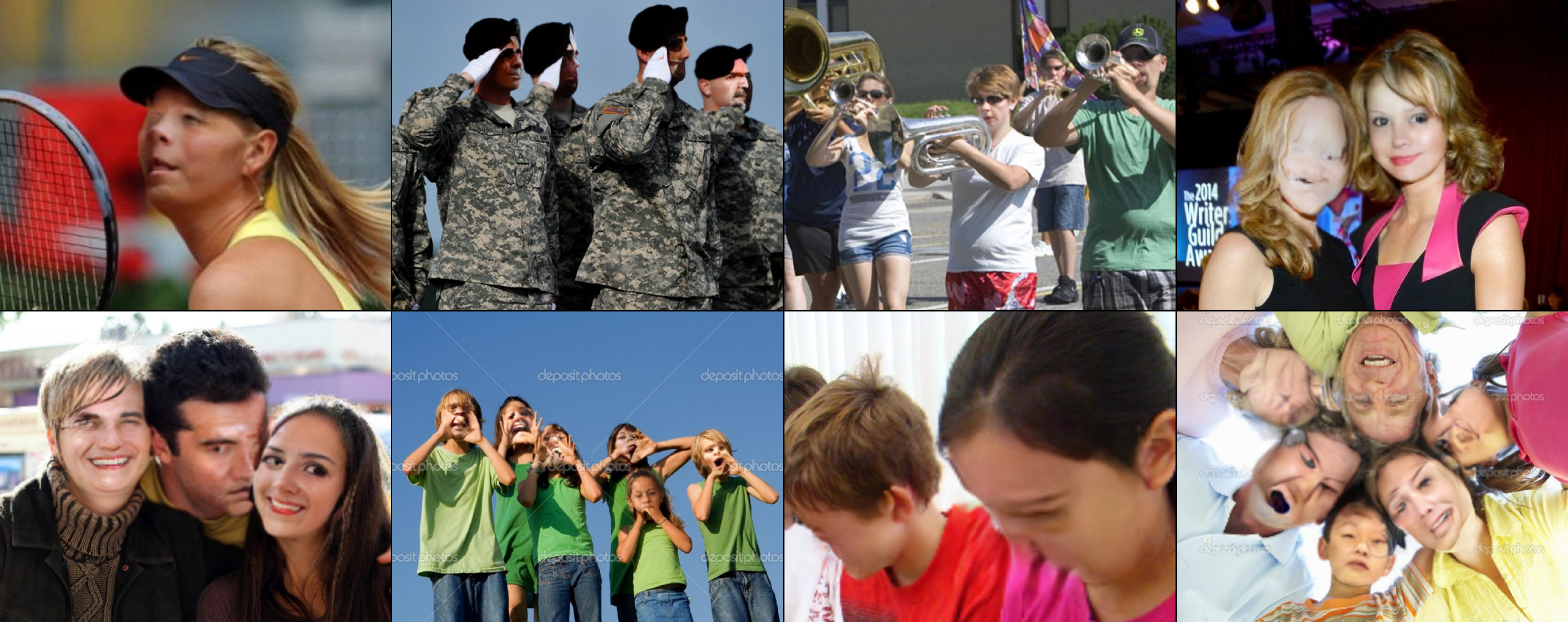}
    \caption{\textbf{Failure Cases of DeepPrivacy} Our proposed solution can generate unrealistic images in cases of high occlusion, difficult background information, and irregular poses. }
    \label{fig:failure_cases}
\end{figure}

Our method proves its ability to generate objectively good images for a diversity of backgrounds and poses. However, it still struggles in several challenging scenarios. \autoref{fig:failure_cases} illustrates some of these. 
These issues can impact the generated image quality, but, by design, our model ensures the removal of all privacy-sensitive information from the face.  

Faces occluded with high fidelity objects are extremely challenging when generating a realistic face.
For example, in \autoref{fig:failure_cases}, several images have persons covering their faces with hands. To generate a face in this scenario requires complex semantic reasoning, which is still a difficult challenge for GANs. 

Handling non-traditional poses can cause our model to generate corrupted faces. We use a sparse pose estimation to describe the face pose, but there is no limitation in our architecture to include a dense pose estimation. A denser pose estimation would, most likely, improve the performance of our model in cases of irregular poses. However, this would set restrictions on the pose estimator and restrict the practical use case of our method. 
\section{Conclusion}


We propose a conditional generative adversarial network, \textit{DeepPrivacy}, to anonymize faces in images without destroying the original data distribution. 
The presented results on the WIDER-Face dataset reflects our model's capability to generate high-quality images. Also, the diversity of images in the WIDER-Face dataset shows the practical applicability of our model. 
The current state-of-the-art face detection method can achieve $99.3\%$ of the original average precision on the anonymized WIDER-Face validation set. In comparison to previous solutions, this is a significant improvement to both the generated image quality and the certainty of anonymization. Furthermore, the presented ablation experiments on the FDF dataset suggests that a larger model size and inclusion of sparse pose information is necessary to generate high-quality images. 

DeepPrivacy is a conceptually simple generative adversarial network, easily extendable for further improvements. Handling irregular poses, difficult occlusions, complex backgrounds, and temporal consistency in videos is still a subject for further work. 
We believe our contribution will be an inspiration for further work into ensuring privacy in visual data.


\section*{Appendix A - Training Details}

We use the same hyperparameters as Karras \etal \cite{Karras2017}, except the following:
We use a batch size of 256, 256, 128, 72 and 48 for resolution 8, 16, 32, 64, and 128.
We use a learning rate of 0.00175 with the Adam optimizer.
For each expansion of the network, we have a transition and stabilization phase of 1.2M images each.
We use an exponential running average for the weights of the generator as this improves overall image quality \cite{yazici2018unusual}.
For the running average, we use a decay $\beta$ given by:

\begin{equation}
\beta = 0.5^{\frac{B}{10^4}},
\end{equation}

where $B$ is the batch size.
Our final model was trained for 17 days on two NVIDIA V100-32GB GPUs. 

\subsection*{Image Pre-Processing}

\begin{figure}
	\centering
	\includegraphics[width=1.0\textwidth]{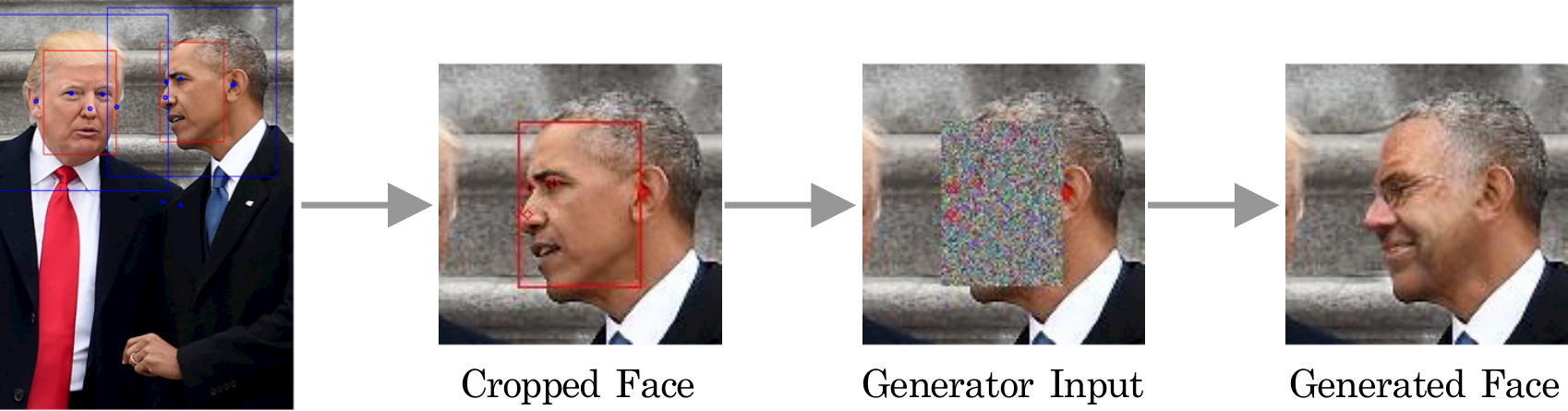}
	\caption{\textbf{Input Pipeline:} Each detected face is cropped to a quadratic image, then we replace the privacy-sensitive information with a constant value, and feed it to the generator. The keypoints are represented as a one-hot encoded image.}
	\label{fig:inpu_pre_processing}
\end{figure}

\autoref{fig:inpu_pre_processing} shows the input pre-processing pipeline.
For each detected face with a bounding box and keypoint detection, we find the smallest possible square bounding box which surrounds the face bounding box. 
Then, we resize the expanded bounding box to the target size ($128 \times 128$).
We replace the pixels within the face bounding box with a constant pixel value of $128$.
Finally, we shift the pixel values to the range $[-1, 1]$.

\subsection*{Tensor Core Modifications}

To utilize tensor cores in NVIDIA's new Volta architecture, we do several modifications to our network, following the requirements of tensor cores.
First, we ensure that each convolutional block use number of filters that are divisible by 8. 
Secondly, we make certain that the batch size for each GPU is divisible by 8.
Further, we use automatic mixed precision for pytorch \cite{nvidia_apex} to significantly improve our training time.
We see an improvement of $220\%$ in terms of training speed with mixed precision training.

\bibliographystyle{splncs04}
\bibliography{egbib}
\end{document}